\documentclass[conference]{IEEEtran}
\IEEEoverridecommandlockouts

\usepackage{amsmath,amssymb}
\usepackage{bm}
\usepackage{graphicx}
\usepackage{xcolor}
\usepackage{cite}
\usepackage{amsthm}
\usepackage{hyperref}
\hypersetup{colorlinks=true, linkcolor=blue, citecolor=blue, urlcolor=blue}

\newcommand*{\tens}[1]{\mathcal{#1}}
\newcommand*{\R}{{\mathbb{R}}}
\newcommand*{\hermconj}{^{\mathsf{H}}}

\newtheorem{theorem}{Theorem}
\newtheorem{definition}{Definition}

\title{A Framework for Directed Hypergraph Signal Processing via tensor t-SVD
\thanks{
C. Mundo-Levano and G. R. Arce are with the Department of Electrical and Computer Engineering,
University of Delaware, Newark, DE 19716, USA.
N. Bello is with the Institute for Financial Services Analytics,
University of Delaware, Newark, DE 19716, USA.
D. L. Lau is with the Department of Electrical and Computer Engineering,
University of Kentucky, Lexington, KY 40506, USA.
This work was partially supported by the National Science Foundation under grants 2230161, 1815992, and 1816003, by the Air Force Office of Scientific Research (AFOSR) under award FA9550-22-1-0362, and by the Institute of Financial Services Analytics, co-sponsored by JP Morgan Chase \& Co.
This work was presented as an oral presentation at the 9th Graph Signal Processing Workshop (GSP 2026), June 8-10, 2026, Madrid, Spain.}
}
\author{
  \IEEEauthorblockN{Carlos Mundo-Levano, Nicol\'{a}s Bello, Daniel L. Lau, Gonzalo R.\ Arce}
}

\begin{document}

\maketitle

\begin{abstract}
We introduce \emph{Directed Hypergraph Signal Processing} (DHGSP), a unified framework that extends graph signal processing to accommodate both higher-order (polyadic) and asymmetric (directional) relationships simultaneously. 
Using the tensor singular value decomposition (t-SVD) within the t-product algebra, we define a novel adjacency tensor for directed hypergraphs, a topologically faithful shift operator, and a lossless Directed Hypergraph Fourier Transform (t-DHGFT).  
Experiments on real traffic networks demonstrate that DHGSP outperforms matrix-based (graph and digraph) and undirected tensor-based (hypergraph) baselines in denoising tasks.
\end{abstract}

\begin{IEEEkeywords}
directed hypergraphs, tensor signal processing, t-SVD, hypergraph Fourier transform, spectral denoising.
\end{IEEEkeywords}

\section{Introduction}
\label{sec:intro}

\noindent
Graph Signal Processing (GSP) extends classical signal processing to data on
irregular, non-Euclidean domains~\cite{shuman2013emerging,ortega2018graph}.
Spectral tools derived from symmetric graph operators enable filtering, sampling,
and frequency analysis on networks~\cite{sandryhaila2013discrete,sandryhaila2014discrete}.
However, two structural features of real-world systems remain outside standard GSP. 
\textbf{Directionality.} Many networks are inherently asymmetric: citations flow
from source to target, nutrients cascade through metabolic pathways, and traffic
propagates along one-way roads. Directed GSP (DGSP) handles these via asymmetric
operators~\cite{marques2020signal}, but asymmetry may yield complex eigenvalues,
non-orthogonal eigenvectors, or non-diagonalizable operators, complicating a
consistent, lossless Fourier transform.
\textbf{Higher-order interactions.} Pairwise edges cannot capture group-level
dependencies such as multi-author collaborations or multi-vehicle intersection
dynamics. Hypergraph Signal Processing (HGSP) allows hyperedges connecting
arbitrary node sets~\cite{zhang2020introducing,pena2023thgsp}, but remains
fundamentally undirected.
No current framework handles \emph{both} simultaneously. Consider information
diffusion where a coalition of influencers collectively targets one individual:
neither a digraph nor an undirected hypergraph can model this.

\begin{figure}[h]
\vspace{-0.3cm}
\centering
\includegraphics[width=0.99\columnwidth]{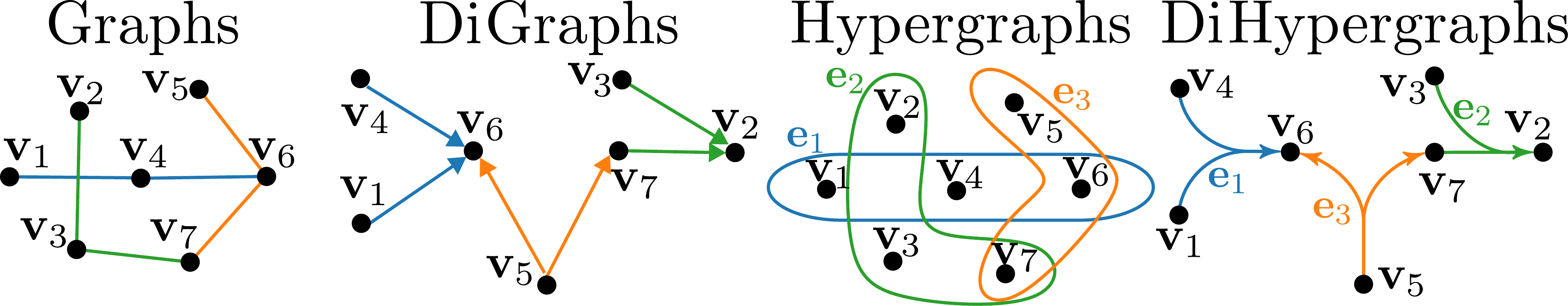}
\vspace{-0.5cm}
\caption{Hierarchy of frameworks: $\mathrm{GSP} \subset \mathrm{DGSP/HGSP} \subset \mathrm{DHGSP}$. Nodes represent data points; edges/hyperarcs encode pairwise, directed-pairwise, undirected-polyadic, or directed-polyadic relationships. DHGSP unifies GSP, DGSP, and HGSP as special cases.}
\label{fig:hierarchy}
\end{figure}

This paper introduces DHGSP, closing this gap with three contributions:
1) A \textbf{tensor adjacency descriptor} built on a canonical B-hyperarc
  decomposition, resolving identifiability and signal cross-talk in prior representations.
2)  A \textbf{lossless t-DHGFT} via t-SVD~\cite{kilmer2011factorization,kilmer2013third}
  of the in-degree Laplacian, guaranteed invertible and Parseval-preserving.
3)  \textbf{Experimental validation} on real traffic data showing superior
  denoising over graph, digraph, and undirected-hypergraph baselines.
DHGSP subsumes GSP, DGSP, and HGSP as special cases (Fig.~\ref{fig:hierarchy}), providing a unified
spectral framework for asymmetric, higher-order networks.

\section{Directed Hypergraph Algebraic Descriptor}
\label{sec:descriptor}
\noindent
A \textbf{directed hypergraph} $\mathrm{DH}=(V,E)$ consists of a vertex set $V$ ($|V|=N$) and a set of \emph{hyperarcs} $E$.  
Each hyperarc $e = (T(e), H(e))$ is an ordered pair of disjoint vertex subsets: the tail (source set) $T(e) \subset V$ and the head (target set) $H(e) \subset V$~\cite{gallo1993directed}.  
A directed graph is the special case $|T(e)|=|H(e)|=1$; an undirected hypergraph results from symmetrizing directions.

\subsection{From Directed Hyperarcs to B-Hyperarcs}

\noindent
Prior tensor representations of directed hypergraphs assign multiple tensor indices to head nodes, causing two fundamental problems~\cite{chen2014circulant,banerjee2018spectrum}: (i) \emph{identifiability}, it is impossible to determine from the tensor alone whether a node is a source or target, and (ii) \emph{signal cross-talk}, shift operations incorrectly mix information among distinct targets.  
The most principled prior approach focuses on Uniform B-hypergraphs (many-to-one hyperarcs)~\cite{gallo2022synchronization}, where the first tensor index exclusively encodes the single head node; but this does not generalize to non-uniform cardinalities.

Our key insight is that any directed hyperarc can be \emph{canonically decomposed} into a collection of B-hyperarcs, one per head node (Fig.~\ref{fig:bhyperarc} in the Appendix).
This decomposition isolates each target's incoming flow, preventing cross-talk by design.

\subsection{Proposed Adjacency Tensor}

\begin{definition}[Adjacency Tensor for Directed Hypergraphs]
Let $\mathrm{DH}$ have $N$ nodes and maximum hyperarc cardinality $M$, and assume the canonical B-hyperarc decomposition, so every hyperarc has a single head.
The adjacency tensor $\tens{A} \in \R^{N^M}$ is built as follows.
For each B-hyperarc $e_i = (T(e_i), \{v_k\})$ with tail set $T(e_i)$ and unique head $v_k$, we set
  $a_{k,\, p_2,\ldots,p_M} = |T(e_i)| / {\alpha_i}$,
where the first index $k$ is reserved for the head node, the remaining indices $\{p_2,\ldots,p_M\}$ range over all length-$(M{-}1)$ permutations (with repetition) of the tail nodes in which each tail appears at least once, and the multinomial normalization factor $\alpha_i$ is the total number of these permutations (given explicitly, with a worked example, in the Appendix; see also Fig.~\ref{fig:tensor_example}).
\end{definition}

\noindent
By reserving the first index exclusively for head nodes and treating each head independently (B-hyperarc decomposition), this definition solves both identifiability and cross-talk.  
At $M=2$, $\tens{A}$ reduces to the standard $N\times N$ adjacency matrix of a directed graph.  
The in-degree and in-Laplacian tensors follow, with $\tens{D}_{\mathrm{in}}$ carrying the in-degrees $d_{\mathrm{in}}(v_k)$ on its superdiagonal:
$
  d_{\mathrm{in}}(v_k) = \sum_{i_2,\ldots,i_M} \tens{A}_{k,i_2,\ldots,i_M},
  \tens{L}_{\mathrm{in}} = \tens{D}_{\mathrm{in}} - \tens{A}.
$

\section{Shift Operator and Topological Localization}
\label{sec:shift}

\noindent
Building on the t-product algebra~\cite{kilmer2011factorization}, the \textbf{directed hypergraph shift} is defined as
  $\vec{\tens{Y}} = \tens{F} * \vec{\tens{X}},$
where $\tens{F}$ is typically $\tens{A}$ or $\tens{L}_{\mathrm{in}}$, and $\vec{\tens{X}} \in \R^{N\times 1 \times N^{M-2}}$ is a tensor signal formed from the underlying node signal $x \in \R^N$ as its $(M{-}1)$-fold outer product, $\vec{\tens{X}} = x \odot \cdots \odot x$ ($\odot$: outer product), encoding polyadic interactions~\cite{pena2023thgsp,zhang2020introducing}.
A symmetrization step~\cite{pena2023thgsp} is applied along the higher (mode $\geq 3$) dimensions to ensure t-product tractability without discarding directional information: it appends a zero slice and reflects entries into a palindrome, yielding $\tens{F}_s \in \R^{N \times N \times N_s^{M-2}}$ with $N_s = 2N+1$.  Because the reflection touches only modes $\geq 3$, the frontal slices are left intact, so the symmetrized shift $\vec{\tens{Y}}_s = \tens{F}_s * \vec{\tens{X}}_s$ preserves all directional structure; the palindromic mode-3 structure renders the DFT slices real, yielding real tubal singular values.
\begin{theorem}[Topological Localization]
Let $\vec{\tens{X}}$ be localized at node $v$ (i.e., $x_v=1$, $x_u=0$ for $u\neq v$).
The first-order shifted signal $\tens{Y}_s^{(1)} = \tens{A}_s * \tens{X}_s$ is non-zero only at nodes $u$ for which $\exists\, e\in E$ with $v \in T(e)$ and $u\in H(e)$.
More generally, after $\ell$ shifts, the signal is non-zero only at nodes reachable from $v$ in exactly $\ell$ directed hyperarc steps.
\end{theorem}

\noindent
The proof follows directly from the B-hyperarc construction and the definition of the t-product. This theorem guarantees that our algebraically defined shift respects the directed topology, a prerequisite for meaningful spectral analysis and causal modeling on directed hypergraphs.

\section{Directed Hypergraph Fourier Transform}
\label{sec:dhgft}
\noindent
The asymmetry of $\tens{L}_{\mathrm{in}}$ (Fig.~\ref{fig:asymmetry} in the Appendix) prevents direct t-eigendecomposition, which requires symmetry for a real orthonormal basis.
We resolve this by extending the self-adjoint dilation technique from directed GSP~\cite{chen2022graph} to tensors.

\subsection{Self-Adjoint Dilation and t-SVD}

\noindent
For $\tens{L}_{\mathrm{in}_s} \in \R^{N\times N \times N_s^{M-2}}$, construct the symmetric dilation
\vspace{-0.4cm}
\begin{equation}
  \tens{S}(\tens{L}_{\mathrm{in}_s}) := \begin{pmatrix} \mathbf{O} & \tens{L}_{\mathrm{in}_s} \\ \tens{L}_{\mathrm{in}_s}\hermconj & \mathbf{O} \end{pmatrix} \in \R^{2N\times 2N \times N_s^{M-2}}.
\vspace{-0.3cm}
\end{equation}

This is symmetric under the t-product by construction, guaranteeing a complete orthonormal eigenbasis with \emph{real} tubal eigenvalues, avoiding the complex spectra of naive directed decompositions.
The t-SVD of $\tens{L}_{\mathrm{in}_s} = \tens{U} * \tens{S} * \tens{V}^H$ (with $\tens{U},\tens{V}$ t-orthogonal and $\tens{S}$ f-diagonal, singular values in ascending order) yields the eigenbasis of the dilation:
\begin{equation}
  \tens{S}(\tens{L}_{\mathrm{in}_s}) = \tens{W} * \begin{pmatrix} \tens{S} & \mathbf{0} \\ \mathbf{0} & -\tens{S} \end{pmatrix} * \tens{W}\hermconj, \,
  \tens{W} = \frac{1}{\sqrt{2}}\begin{pmatrix} \tens{U} & \tens{U} \\ \tens{V} & -\tens{V} \end{pmatrix}.
\end{equation}

\subsection{t-DHGFT and Inverse}

Embedding $\tens{X}_s$ isometrically into the dilated space as $\bm{\tens{X}}_s = \frac{1}{\sqrt{2}}(\tens{I}_s;\, \tens{I}_s)\hermconj * \tens{X}_s$ (with $\tens{I}_s$ the symmetrized identity tensor; the embedding preserves energy for Parseval's identity), we define the \textbf{t-DHGFT}:
\begin{equation}
  \vec{\tens{X}}_{F_s} = \tens{W}\hermconj * \bm{\tens{X}}_s = \frac{1}{2}\begin{pmatrix} (\tens{U}\hermconj + \tens{V}\hermconj)*\tens{X}_s \\ (\tens{U}\hermconj - \tens{V}\hermconj)*\tens{X}_s \end{pmatrix},
\end{equation}
and the \textbf{inverse t-DHGFT}:
\begin{equation}
  \vec{\tens{X}}_s = \frac{1}{2}\Bigl(\tens{U}*(\vec{\tens{X}}_{F_{s,1}}+\vec{\tens{X}}_{F_{s,2}}) + \tens{V}*(\vec{\tens{X}}_{F_{s,1}}-\vec{\tens{X}}_{F_{s,2}})\Bigr).
\end{equation}
The orthogonality of $\tens{W}$ guarantees perfect reconstruction and Parseval’s identity. Frequency is ordered by the \emph{directed hypergraph total variation} $\mathrm{TV}(\vec{\tens{X}}_s) = \|\tens{L}_{\mathrm{in}}*\vec{\tens{X}}_s\|^2$: for a right singular vector $\vec{\tens{V}}_j$, $\mathrm{TV}(\vec{\tens{V}}_j) = \sigma_j^2$, so the singular values $\sigma_1 \leq \sigma_2 \leq \cdots \leq \sigma_{N_s}$ order the modes from smooth (small $\sigma_j$) to rapidly oscillating (large $\sigma_j$; see Fig.~\ref{fig:directed_hypergraph_frequency_ordering_real} in the Appendix).

\section{Spectral Denoising Experiments}
\label{sec:experiments}

We validate DHGSP via spectral denoising on a real traffic dataset (Macheng, China: 153 nodes)~\cite{gsp_traffic}, comparing against (i) undirected GSP (symmetric Laplacian), (ii) DGSP (SVD of directed Laplacian), and (iii) undirected HGSP via t-product~\cite{pena2023thgsp}.
Directed hypergraphs are constructed by grouping directed edges into hyperarcs following~\cite{zhao2024dhmconv}.
\textbf{Method.} The signal is corrupted with additive white Gaussian noise and reconstructed via low-pass hard-thresholding in the t-DHGFT domain: only the $K$ lowest-frequency components are retained. The cutoff $K$ is varied across the full spectrum, following the protocol in \cite{pena2023thgsp}.
\textbf{Results.} Fig.~\ref{fig:denoising} in the Appendix shows MAE vs.\ conserved frequencies on the Macheng dataset.
DHGSP consistently achieves lower MAE than all baselines across the full frequency range, with the improvement most pronounced at intermediate bandwidths (50--60\% of spectrum).
The gain over undirected HGSP confirms that directionality provides informative structure beyond higher-order topology alone; the gain over DGSP confirms that polyadic interactions capture dependencies missed by pairwise edges.

\section{Conclusion}
\label{sec:conclusion}

We have introduced Directed Hypergraph Signal Processing (DHGSP), a unified framework built on the t-SVD algebra that models both directional and higher-order relationships.  
Three core contributions: a semantically faithful adjacency tensor, a topologically localizing shift operator, and a lossless Fourier transform via self-adjoint dilation, constitute a coherent spectral theory for directed hypergraphs.  
Experimental results on real traffic networks validate the framework's advantage over graph, digraph, and undirected hypergraph baselines.

Future work includes: sampling theory for directed hypergraphs, compression bounds, spectral clustering algorithms, and directed hypergraph neural networks based on learned t-product filter banks.

\bibliographystyle{IEEEtran}
\bibliography{bibliography}

\newpage
\appendix
\section*{Appendix: Figures and Tables}

\begin{figure}[h]
\centering
\includegraphics[width=\columnwidth]{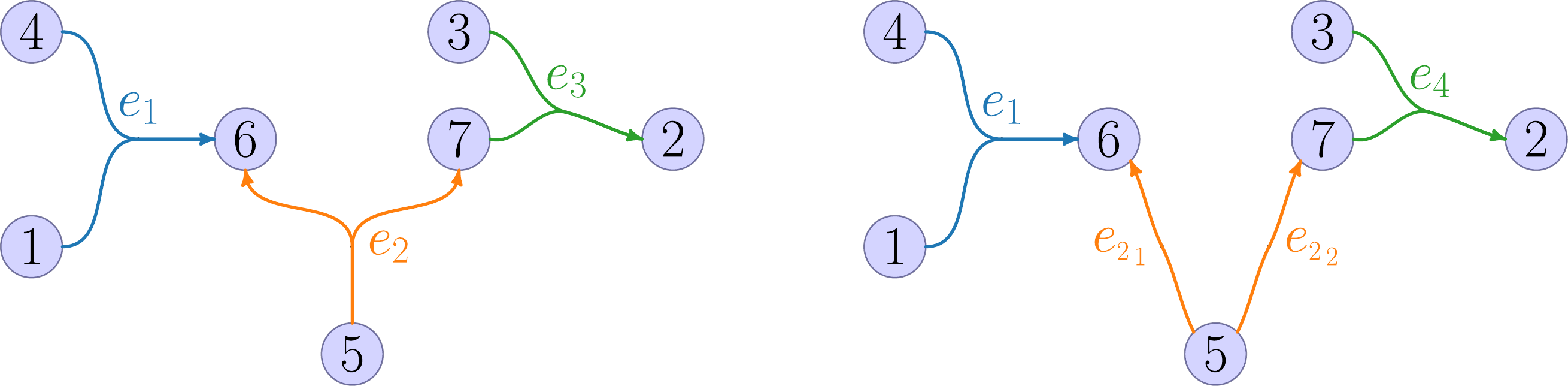}
\caption{Decomposition of a many-to-many hyperarc (left) into a collection of B-hyperarcs, one per head node (right). This canonical decomposition prevents signal cross-talk between targets (e.g., $v_6$ and $v_7$).}
\label{fig:bhyperarc}
\end{figure}

\begin{figure}[h]
\centering
\includegraphics[width=0.7\columnwidth]{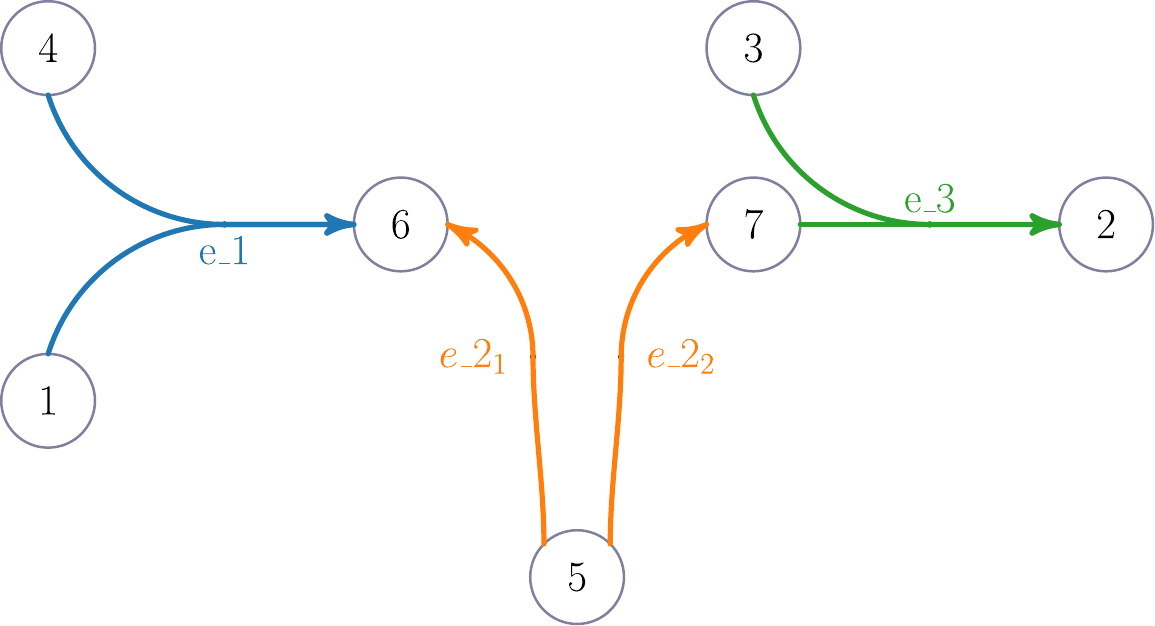}
\caption{Worked example of the adjacency tensor for $M=3$, so $\tens{A}\in\R^{N\times N\times N}$ with indices $(i_1,i_2,i_3)$ and $i_1$ reserved for the head node. For the hyperarcs shown: $e_1\!:\{v_1,v_4\}\!\to\!\{v_6\}$ gives $a_{6,1,4}=a_{6,4,1}=1$; $e_{2_1}\!:\{v_5\}\!\to\!\{v_6\}$ gives $a_{6,5,5}=1$; $e_{2_2}\!:\{v_5\}\!\to\!\{v_7\}$ gives $a_{7,5,5}=1$; $e_3\!:\{v_3,v_7\}\!\to\!\{v_2\}$ gives $a_{2,3,7}=a_{2,7,3}=1$. The in-degree of $v_6$ is $d_{6,6,6}=a_{6,1,4}+a_{6,4,1}+a_{6,5,5}=3$.}
\label{fig:tensor_example}
\end{figure}

The normalization $\alpha_i$ in the Definition is the total number of ordered length-$(M{-}1)$ permutations of the $c_i = |T(e_i)|$ distinct tail nodes in which each tail appears at least once:
\begin{equation*}
  \alpha_i = \sum_{\substack{k_2,\ldots,k_{c_i} \geq 1 \\ k_2+\cdots+k_{c_i} = M-1}}
  \frac{(M-1)!}{k_2!\cdots k_{c_i}!}.
\end{equation*}
For example, at $M=3$: a hyperarc with $|T|=2$ gives $\alpha = 2!/(1!\,1!) = 2$, so $a = |T|/\alpha = 1$; a hyperarc with $|T|=1$ gives $\alpha = 2!/2! = 1$, so $a = 1$.

\begin{figure}[h]
\centering
\includegraphics[width=0.7\columnwidth]{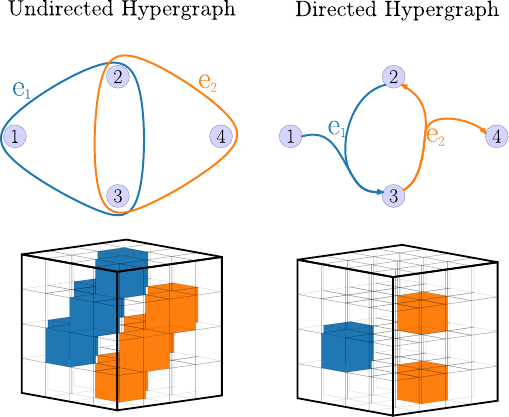}
\caption{Frontal slices of the adjacency tensor: undirected hypergraph (left, symmetric) versus directed hypergraph (right, asymmetric). The asymmetry requires the t-SVD and dilation operators rather than a simple t-eigendecomposition.}
\label{fig:asymmetry}
\end{figure}

\begin{figure}[h]
\begin{center}
\includegraphics[width=0.99\linewidth]{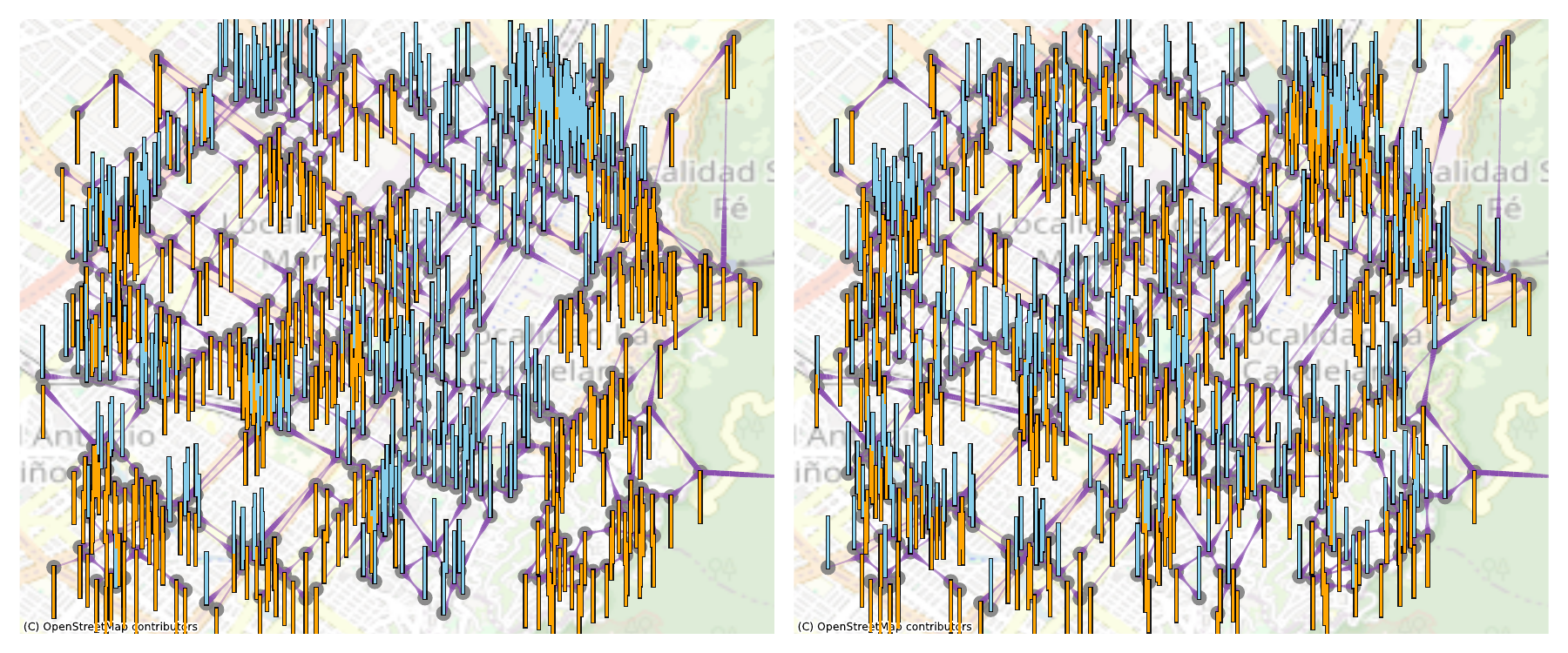}
\end{center}
\vspace{-0.25in}
\caption{First frontal slices of the eigenmatrices for traffic data from Bogotá, Colombia (689 nodes)~\cite{gsp_traffic}, modeled as a directed hypergraph by grouping edges into hyperarcs~\cite{zhao2024dhmconv}. Left: low-frequency modes ($\sim$5\% of the spectrum); Right: high-frequency modes ($\sim$95\% of the spectrum). Higher frequencies exhibit substantially greater spatial variation.}
\label{fig:directed_hypergraph_frequency_ordering_real}
\vspace{0.3cm}
\end{figure}

\begin{figure}[h]
\centering
\includegraphics[width=0.8\columnwidth]{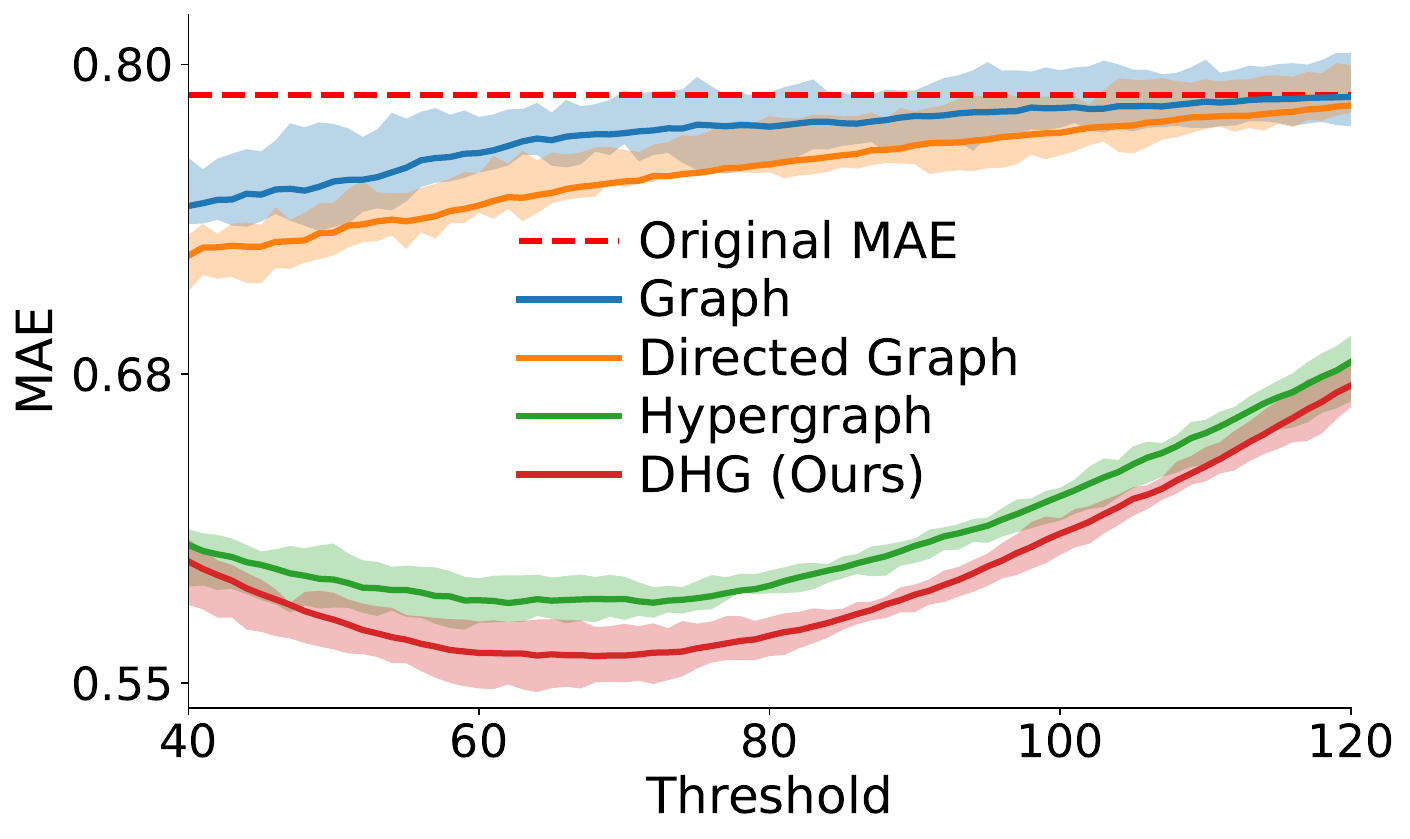}
\caption{Mean absolute error (MAE) versus total conserved frequencies for spectral denoising on Macheng, China traffic data (153 nodes)~\cite{gsp_traffic}. DHGSP (red) consistently outperforms GSP (blue), DGSP (orange), and HGSP (green) across all bandwidths. Shaded region shows the 40th--60th percentile over 100 trials.}
\label{fig:denoising}
\end{figure}

\end{document}